\documentclass[a4paper,fleqn]{cas-dc}

\usepackage[authoryear]{natbib}

\usepackage{changes} 

\usepackage{colortbl}  
\definecolor{Gray}{gray}{0.94}
\newcommand{\shline}{\specialrule{.05em}{.05em}{.05em}}

\def\tsc#1{\csdef{#1}{\textsc{\lowercase{#1}}\xspace}}
\tsc{WGM}
\tsc{QE}
\tsc{EP}
\tsc{PMS}
\tsc{BEC}
\tsc{DE}

\begin{document}
\let\WriteBookmarks\relax
\def\floatpagepagefraction{1}
\def\textpagefraction{.001}

\shorttitle{}

\shortauthors{Zefeng Qian et~al}

\title [mode = title]{Joint Image-Instance Spatial-Temporal Attention for Few-shot Action Recognition}                      



%
\author[a]{Zefeng Qian}[type=editor,
                        style=chinese, 
                        ]


\ead{zefeng_qian@sjtu.edu.cn}

\author[a]{Chongyang Zhang}[style=chines]
\cormark[1]
\ead{sunny_zhang@sjtu.edu.cn}




\author[b]{Yifei Huang}[style=chinese]
\ead{hyf@iis.u-tokyo.ac.jp}

\author[c]{Gang Wang}[style=chinese]

\author[c]{Jiangyong Ying}[style=chinese]


\affiliation[a]{organization={Shanghai Jiao Tong University},
    city={Shanghai},
    country={China}}
\affiliation[b]{organization={The University of Tokyo},
    city={Tokyo},
    country={Japan}} 
\affiliation[c]{organization={E-surfing Vision Technology Co., Ltd.},
    city={Hangzhou},
    country={China}}

\cortext[cor1]{Corresponding author}

\begin{abstract}
Few-shot Action Recognition (FSAR) constitutes a crucial challenge in computer vision, entailing the recognition of actions from a limited set of examples. Recent approaches mainly focus on employing image-level features to construct temporal dependencies and generate prototypes for each action category. 
However, a considerable number of these methods utilize mainly image-level features that incorporate background noise and focus insufficiently on real foreground (action-related instances), thereby compromising the recognition capability, particularly in the few-shot scenario. 
To tackle this issue, we propose a novel joint Image-Instance level Spatial-temporal attention approach (I$^2$ST) for Few-shot Action Recognition. The core concept of I$^2$ST is to perceive the action-related instances and integrate them with image features via spatial-temporal attention. 
Specifically, I$^2$ST consists of two key components: Action-related Instance Perception and Joint Image-Instance Spatial-temporal Attention.
Given the basic representations from the feature extractor, the Action-related Instance Perception is introduced to perceive action-related instances under the guidance of a text-guided segmentation model. Subsequently, the Joint Image-Instance Spatial-temporal Attention is used to construct the feature dependency between instances and images.
To enhance the prototype representations of different categories of videos, a pair of spatial-temporal attention sub-modules is introduced to combine image features and instance embeddings across both temporal and spatial dimensions, and a global fusion sub-module is utilized to aggregate global contextual information, then robust action video prototypes can be formed.
Finally, based on the video prototype, a Global-Local Prototype Matching is performed for reliable few-shot video matching. In this manner, our proposed I$^2$ST can effectively exploit the foreground instance-level cues and model more accurate spatial-temporal relationships for the complex few-shot video recognition scenarios. Extensive experiments across standard few-shot benchmarks demonstrate that the proposed framework outperforms existing methods and achieves state-of-the-art performance under various few-shot settings.

\end{abstract}



\maketitle

\section{Introduction}

Few-shot learning \citep{Bart_Ullman_2005, One-shot2006} aims to develop a model that can generalize to novel categories using only a limited number of labeled samples.
This task is particularly challenging because directly fine-tuning deep learning models on limited labeled data can easily lead to overfitting, significantly hindering their generalization ability.
In this paper, we study the problem of Few-shot Action Recognition (FSAR), which is tasked with classifying unseen action classes with extremely few annotated video examples.
Compared to image-based tasks \citep{One-shot2006, hariharan2017low, li2020adversarial, schwartz2018delta-encoder, wang2018low, metagan}, videos contain richer spatial-temporal and contextual information, which requires not only a deep understanding of contextually relevant semantics but also the extensive modeling of spatial-temporal relationships.
Consequently, it is more complex and challenging for deep learning algorithms to learn and interpret effectively.

\begin{figure}[t]
  \centering
   \includegraphics[width=\linewidth]{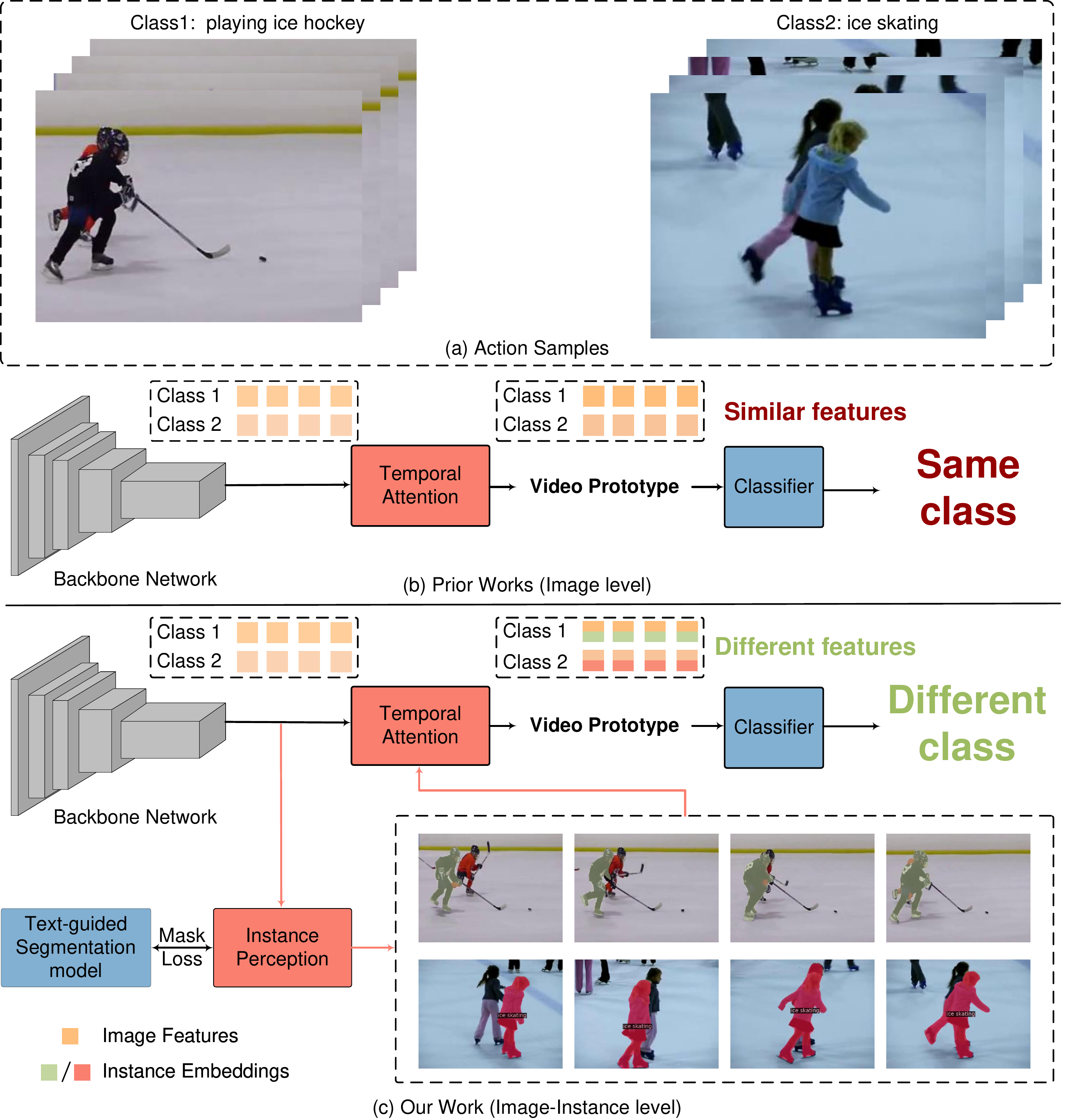}
   \caption{
   Illustration of our motivation. 
   (a) Visualization of two similar actions: "playing ice hockey" and "ice skating". (b) Framework of prior works (Image Level). (c) Our Framework (Image-Instance Level). Compared to (b), our framework (c) explicitly perceives the action-related instances from the image and merges them via spatial-temporal Attention, generating more discriminative prototypes.
   }
   \label{fig:Motivation}
\end{figure}

To solve the data-scarcity problem in FSAR, most previous FSAR methods \citep{2018-CMN, 2020-CMN, 2020-OTAM, perrett2021trx, ARN-ECCV, 2022-HyRSM} follow a "metric-based meta-learning" paradigm, where a common embedding space is first learned through episodic training, and an explicit or implicit alignment metric is employed to calculate the similarities between the query video and support videos for classification in an episodic task.
Typically, OTAM \citep{2020-OTAM} leverages a deep network to extract video features and explicitly estimates an ordered temporal alignment path to match the frames of two videos. TRX \citep{perrett2021trx} models temporal relations by representing the video as tuples consisting of a few sparse frames, and compares the similarity of the query-support pair in a part-based manner. 
Although these methods have achieved remarkable performance, most of them consider mainly image-level features to model temporal dependencies, often overlooking action-related instance information.
However, in some cases, utilizing only image-level features may be less effective for action recognition and limit the generalization performance of the model. 
For example, as shown in Figure \ref{fig:Motivation} (a), the two action classes "playing hockey" and "ice skating" are very similar in their background information. The critical distinction lies in the action-related instance——whether the subject in the video is playing ice hockey or just skating. 
These two action classes are difficult to distinguish due to their similar image-level feature representations, which contain excessive similar background information and insufficient action-related instance information, as shown in Figure \ref{fig:Motivation} (b).

{To alleviate this, several studies have explored the incorporation of instance-level information to enhance video representations. For example, SR-CNNs \cite{wang2016sr-cnns} explicitly leverage semantic cues from pre-trained object detectors to improve action understanding. Similarly, Chen \textit{et al.} \citep{chen2023video} introduce a multi-region attention module for perception the region-aware visual features beyond the original global feature. 
Inspired by this, recent research in FSAR has sought to enhance image-level feature representations by exploiting local features or low-level features in FSAR.
} 
For instance, HCL \citep{HCL} enhance the discrimination of the prototype global, temporal, and spatial levels feature.
SloshNet \citep{sloshnet} exploits the fusion structure of the low-level and high-level spatial features in FSAR. 
Although these methods enhance the action prototype with combining local patch features and image-level global representations, they still fail to explicitly capture action-related instances, resulting in limited performance improvements.
Further more, to explicitly capture foreground object features, Huang \textit{et al.} \citep{huang2022compound, huang2024matching} employs an extra object detector \citep{mask_RCNN} to extract object features in addition to the image-level features, and encode spatial-temporal relation with the compound prototypes. 
However, object detectors can still be influenced especially when the background is cluttered, since they have no knowledge about which one is the action-related object.

In this work, we explicitly consider the foreground action-related instances, and propose a novel joint Image-Instance level Spatial-Temporal attention model for Few-shot Action Recognition, namely I$^2$ST, as shown in Figure \ref{fig:Motivation} (c).
To explicitly capture the foreground action-related instances, we employ an Instance Perception Module.
We ensure that the instance embeddings capture precise foreground information by designing this module as an encoder-decoder structure, where the decoder is guided by a text-guided segmentation model.
To effectively enhance the video representation, we comprehensively integrate instance-level and image-level information with Joint Image-Instance Spatial-Temporal Attention.
More specifically, to enhance the image-level representations with action-related instance embeddings, a self-attention module is employed separately along the spatial and temporal dimensions.
Then, we employed a global fusion sub-module, which introduces a learnable token to concatenate with these features, to perform multi-head attention for global contextual information aggregation and robust prototype construction.
Finally, built upon the video prototype, a Global-Local Prototype Matching is performed for reliable few-shot video matching. Concretely, the video prototype consists of a global prototype and a local prototype, where the global prototype captures the action information from the entire video, and each local prototype focuses on a specific temporal location of the video. The final similarity score is calculated as the average similarity between the global prototype and local prototypes between the query and support videos.
In this way, our proposed I$^2$ST can effectively enhance the video representation with foreground instance-level cues, and improve performance in FSAR.

Experimental results on multiple widely used benchmarks demonstrate that our I$^2$ST outperforms other advanced FSAR methods and achieves state-of-the-art performance. We also perform multiple ablation studies and qualitative visualizations, all of which prove the effectiveness of both our Instance Perception Module and our Spatial-temporal Attention Module.
In summary, our contributions are listed as follows:
\begin{itemize}
\itemsep=0pt
\item We propose a novel method I$^2$ST for Few-shot Action Recognition to effectively leverage both instance-level and image-level information. To the best of our knowledge, this is the first work to consider instance-level cues without an extra detector in Few-shot Action Recognition.
\item We employ Action-related Instance Perception to capture foreground action-related instance information, and Joint Image-Instance Spatial-Temporal Attention to integrate instance-level and image-level information across spatial and temporal dimensions.
\item We conduct extensive experiments across five widely-used benchmarks to validate the effectiveness of the proposed I$^2$ST. The results show that our method achieves state-of-the-art performance.
\end{itemize}

\section{Related Work}
In this section, we will briefly review the work closely related to this work, mainly including few-shot learning, few-shot action recognition (FSAR) and foreground perception based FSAR.

\subsection{Few-shot Learning}

Few-shot learning \citep{One-shot2006} aims to identify new concepts with just a few labeled training instances.
Mainstream approaches to few-shot learning can be categorized into three types: augmentation-based, gradient optimization, and metric-based methods.
Augmentation-based methods \citep{chen2019image,wang2018low, schwartz2018delta-encoder, hariharan2017low, metagan, li2020adversarial} typically focus on generating additional samples to mitigate the issue of data scarcity.
For example, Wang \textit{et al.} \citep{wang2018low} explored the integration of meta-learning with an innovative "hallucinator" that produces additional training samples, significantly improving low-shot learning performance.
MetaGAN \citep{metagan} employs a task-conditioned adversarial generator to distinguish between real and fake data, enabling few-shot classifiers to create finer decision boundaries for better generalization.
Gradient optimization methods \citep{MAML,jamal2019task, li2017meta-sgd,rajeswaran2019meta, ravi2016optimization, rusu2018meta} are designed to refine the network's optimization process, enabling rapid adjustment of the model towards an optimal point.
Metric-based methods create a unified metric space for both known and new classes, simplifying the comparison of query and support samples by using distances within this metric space.
These methods commonly use few-shot matching to classify query samples based on metrics like Euclidean distance \citep{oreshkin2018tadam, snell2017prototypical, wang2019simpleshot, zhang2020tapnet, wu2022superclass, mai2019attentive}, cosine distance \citep{qiao2018few, vinyals2016matchingNet, cao2021few}, and learnable metrics \citep{hao2019collect, sung2018learning, wang2020cooperative}.
Our method also follows metric-based methodologies, focusing on the perception of action-related instances and joint spatial-temporal attention, thereby improving the accuracy of Few-shot Action Recognition.

\subsection{Few-shot Action Recognition}

Few-shot Action Recognition (FSAR) is a sub-field of few-shot learning that deals with videos, which contain complex temporal information.
Existing Few-shot Action Recognition methods \citep{2018-CMN, 2020-CMN, 2020-OTAM,zhu2021few} mainly follow the metric-based meta-learning paradigm \citep{snell2017prototypical} to optimize the model and design robust alignment metrics to calculate the distances between the query and support samples for classification, due to its simplicity and effectiveness. 
To exploit the temporal cues, many approaches focus on local frame-level (or segment-level) alignment between query and support videos.
Notably, OTAM \citep{2020-OTAM} designs a different dynamic time warping \citep{muller2007dynamic} algorithm to temporally align two video sequences. 
TARN \citep{TARN} and MPRE \citep{MPRE} both introduce temporal attention mechanisms to help the network generalize better to unseen classes.
TRX \citep{perrett2021trx} adopts an attention mechanism to align each query sub-sequence against all sub-sequences in the support set and then fuses their matching scores for classification. 
HyRSM \citep{2022-HyRSM}, MTFAN \citep{MTFAN} and GgHM \citep{xing2023GgHM} emphasize the importance of learning task-specific features in Few-shot Action Recognition.

Furthermore, to enhance the representational capacity of features, some methods \citep{luo2021dense,han2022modeling,xiang2023generative, su2025semantic} exploit the complementarity of different modalities in video-related tasks. In particular, mainstream approaches \cite{VD-ZSAR, ResT, ETSAN, DeCalGAN, gao2023learning} in zero-shot action recognition achieve knowledge transfer from seen to unseen actions by embedding videos and action classes into a shared embedding space. 
Inspired by this, SRPN \citep{wang2021semantic} trains a generative model to imitate the semantic labels and fuses the visual features and the generated textual features by element-wise addition.
Recently, several methods \citep{wang2023clip, xing2023ma-clip, CLIP-CPM2C} propose to generate reliable prototypes with the guidance textual semantic priors in CLIP.
Although our method utilizes text semantic information, it only uses text information during the training phase to explicitly guide the learning of instance embeddings. It's text-free during the inference phase. Furthermore, unlike the following methods, we do not use the CLIP-RN50 backbone \citep{radford2021clip}, whose training data may include Few-shot Action Recognition benchmark datasets.

\begin{figure*}[t]
  \centering
  \includegraphics[width=\textwidth]{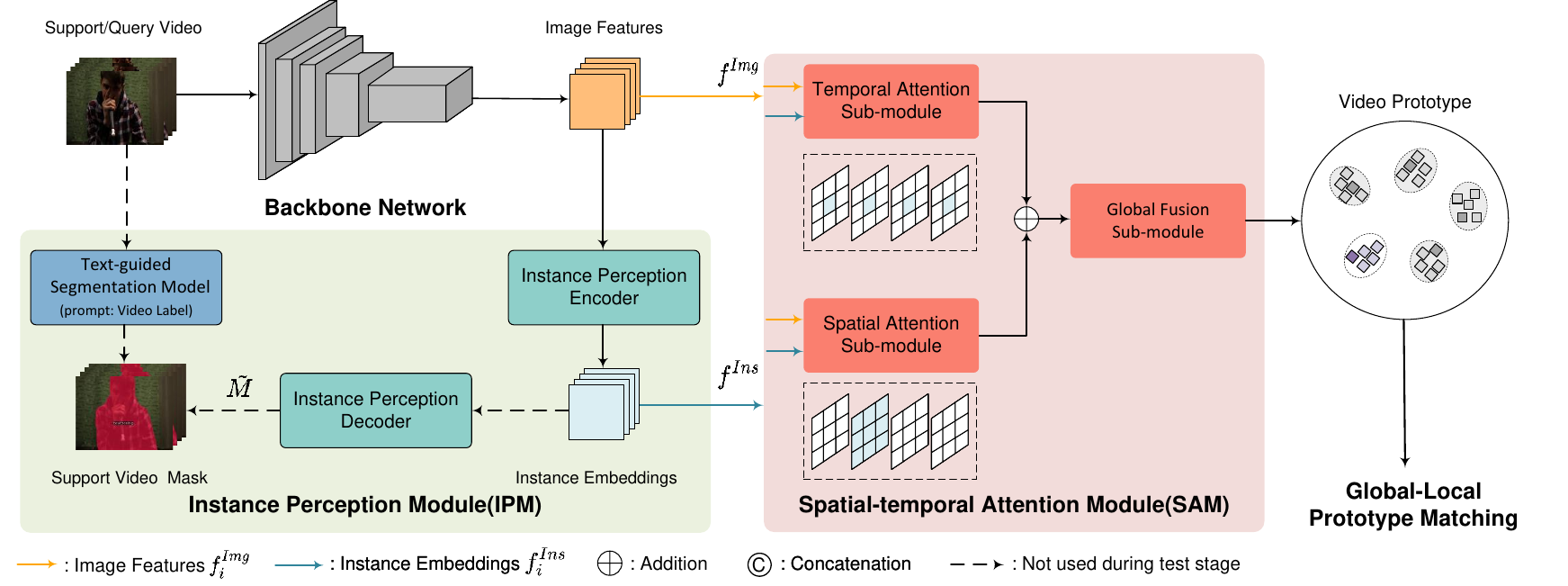}
   \caption{Overview of the proposed I$^2$ST. Given the support and query videos, we first utilize a feature extractor to encode the features of the images in the video.
   We then feed these image features into the instance perception module to extract instance embeddings and recover the action-related instance mask (the dashed line indicates that it was only used during the training phase).
   Then, the image features and the instance embeddings are fed into a Spatial-temporal Attention Module to merge foreground and background information of action videos across both temporal and spatial dimensions.
   Finally, the query video is classified based on the global-local matching results of the obtained video prototypes.
   For the convenience of illustration, other videos involved in a few-shot task are omitted from the figure.}
   \label{fig:framework}
\end{figure*}

\subsection{Foreground Perception based FASR}
{
Previous studies \citep{wang2016sr-cnns, fang2018pairwise, zhou2023can, chen2023video, su2024vpe} have demonstrated that incorporating instance-level information into action recognition can effectively enhance the discrimination of action features and improve model performance. 
Inspired by this, several studies have started to concentrate on enhancing the video presentation with instance-level information in FSAR.
These approaches can generally be divided into two main types: hierarchical methods and detector-based methods.
Hierarchical methods improve action feature representations by adaptively integrating multi-scale features.
}
For example, HCL \citep{HCL} introduces a hierarchical matching model to support comprehensive similarity measures at the global, temporal, and spatial levels via a zoom-in matching module. 
SloshNet \citep{sloshnet} exploits the automatic search for the best fusion structure of the low-level and high-level spatial features in different scenarios with a feature fusion architecture search module.
TADRNet \citep{tadrnet} utilizes a fine-grained local representation network to boost the representation ability of local features of samples and compensate for the weakness of only utilizing the global characteristics of samples for prediction to a certain extent. 
SA-CT \citep{SA-CT} proposes a novel spatial cross-attention module to model the spatial relation, handling the spatial misalignment between two videos in FSAR.
HCR \citep{hcr} design a Parts Attention Module to pay attention to pre-defined human body parts and other action-relevant cues.
Although these methods enrich image-level feature representations with spatial context information and achieve competitive results, they do not explicitly encode the foreground action-related instances, resulting in limited performance improvements.

{Furthermore, detector-based methods introduce an external object detector into FSAR to extract object features alongside image-level representations.}
For instance, to accurately capture object features, Huang \textit{et al.} \citep{huang2022compound, huang2024matching} employs an extra object detector \citep{mask_RCNN} to extract object features in addition to the image-level features, and encode spatial-temporal relation with the compound prototypes. 
Similarly, MGTSN \citep{mgtsn} utilizes a general segmentation model to obtain the mask sequences and proposes a two-stream feature fusion module to fusion the features of the RGB and segmentation mask frame sequences.
However, the general object detector or segmentation model can still be influenced, especially when the background is cluttered, as they have no knowledge about which one is the action-related object or instance. 
{
Moreover, the use of additional detectors introduces extra computational overhead during inference.
Additionally, the integration of such detectors introduces significant computational overhead during inference.
In this paper, we propose a novel Joint Image-Instance Spatial-Temporal attention model, which explicitly extracts foreground action-related instance information in FSAR without introducing any additional detector model during inference.
}

\section{Method}
In this section, we first describe the problem formulation of Few-shot Action Recognition. Then, we present the architecture overview of the I$^2$ST and detail the important components, including the Action-related Instance Perception, the Joint Image-Instance Spatial-Temporal Attention, and the Global-Local Prototype Matching.

\subsection{Problem Formulation}
The goal of Few-shot Action Recognition (FSAR) is to classify an unlabelled query video into one of the action classes in the 'support set'.
Given a training set $\mathcal{D}_{train}=\{(v_{i},y_{i}),y_{i}\in \mathcal{C}_{train}\}$ and a test set $\mathcal{D}_{test}=\{(v_{i},y_{i}),y_{i}\in \mathcal{C}_{test}\}$, we train a deep network on the training set and then verify the generalization of the trained model on the test set ($\mathcal{C}_{train}\cap \mathcal{C}_{test}=\emptyset$). 
To make training more faithful to the test environment, we adopt the episode-based meta-learning strategy \citep{vinyals2016matchingNet} to optimize the network, as in previous approaches \citep{vinyals2016matchingNet, 2020-OTAM, perrett2021trx,2023-molo}.
In each episode, there is a support set $\mathcal{S}$ containing $N$ classes and $K$ videos per class (called the $N$-way $K$-shot task) and a query set $\mathcal{Q}$ containing query samples to be classified. 
The optimization objective is to categorize the query videos in $\mathcal{Q}$ into the correct class among $N$ classes based on feature distance metrics.
During inference, a large number of episodes are randomly selected from the test set, and the average accuracy is utilized to evaluate the few-shot performance of the trained model.

\begin{figure}[t]
  \centering
   \includegraphics[width=0.95\linewidth]{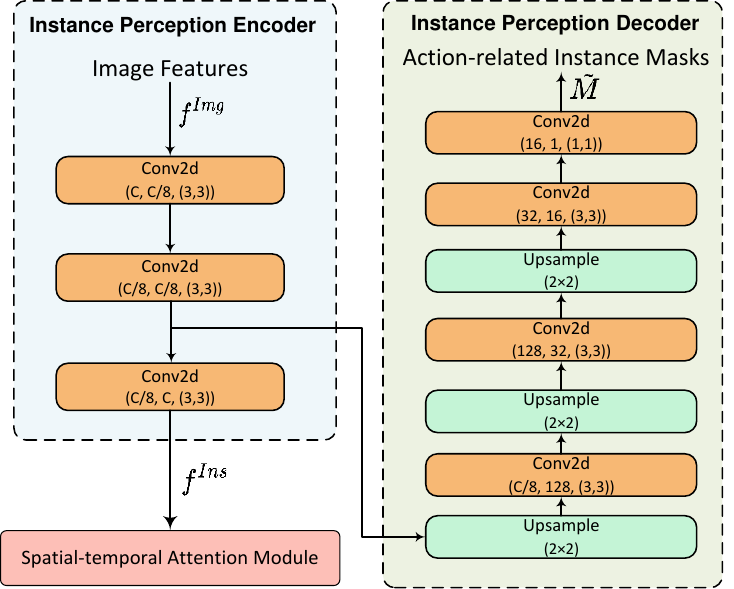}
   \caption{The architecture details the proposed Instance Perception Module (IPM), including an Instance Perception Encoder and an Instance Perception Decoder.
   }
   \label{fig:autodecoder}
\end{figure}

\subsection{Architecture Overview}
An overview of our framework is provided in Figure \ref{fig:framework}, which mainly consists of three components: an Instance Perception Module (IPM), a Spatial-temporal Attention Module (SAM), and a Global-Local Prototype Matching.
Following previous methods \citep{2020-OTAM,perrett2021trx,2022-HyRSM,2023-molo}, we first divide it into $T$ segments and extract a snippet from each segment and utilize the feature extractor (\textit{i.e.} ResNet-50 \citep{2016-resnet}) to generate the image features of support samples and query samples. 
Then, to explicitly perceive foreground action-related instances, we propose a novel Instance Perception Module (IPM) to obtain the instance embeddings from image features. 
The IPM employs a lightweight encoder-decoder structure to capture instance embeddings from complex scenarios with the guidance of a text-guided segmentation model.
Subsequently, the Spatial-temporal Attention Module (SAM) integrates instance-level and image-level information to generate robust and discriminative action prototypes for matching. 
Specifically, the spatial and temporal attention sub-modules integrate these features to generate spatial and temporal fusion features, respectively.
After that, the global fusion module merge them and generate the action prototype with multi-head attention.
Finally, we employ a Global-Local Prototype Matching to calculate similarity for query video classification. We introduce the details of each component below.

\subsection{Action-related Instance Perception}
Most FSAR methods overlook the critical instance-level information in samples, which represents the essential characteristics of actions. To address this limitation, we propose a novel Instance Perception Module (IPM) that explicitly perceives foreground action-related instances to enhance image-level features, enabling more accurate and robust action recognition.
This module is composed of two key components: the Instance Perception Encoder and the Instance Perception Decoder, as shown in Figure \ref{fig:autodecoder}.
To capture precise action-related instances information, the module adopts an encoder-decoder structure, with the decoder guided by a pre-trained text-guided segmentation model \citep{SEEM}.

Specifically, the image features $f^{Img}\in \mathbb{R}^{T \times C \times H \times W}$  are first generated by the feature extractor (without using the last average pooling operation and the fully connected layer), Where $T$ is the number of sparsely sampled video frames and $C$ represent channel dimensions. Then, we utilize a streamlined and lightweight Instance Perception Encoder to efficiently capture action-related instance embeddings $f^{Ins}\in  \mathbb{R}^{T \times C \times H \times W}$, which consist of $2D$ spatial convolution layers.
The formulation is expressed as follows:
\begin{equation}
     f^{Ins}= \mathbf{IPE}(f^{Img})
\end{equation}
where $\mathbf{IPE}$ means the proposed Instance Perception Encoder.

To explicitly guide the perception of instance embeddings, we employ an Instance Perception Decoder to accurately reconstruct action-related instance masks, which can be written as:
\begin{equation}
     \Tilde{M}= \mathbf{IPD}(f^{Ins})
\end{equation}
where $\Tilde{M}$ denotes the output action-related instance masks. The $\mathbf{IPD}$ refers to the Instance Perception Decoder, which is composed of a set of $2D$ spatial convolution layers and upsampling layers, as shown in Figure \ref{fig:autodecoder}. 
We leverage a pre-trained text-guided segmentation model \citep{SEEM}, inputting the action video clip and taking its corresponding semantic label as a prompt to generate the  action-related instance masks $M$.
With the action-related instance masks $M$, we employ an $\mathbb{L}_2$ loss function to guide the network in accurately reconstructing action-related instance masks, thereby facilitating the extraction of effective instance embeddings, which can be written as:
\begin{equation}
    \mathcal{L}_{Mask} = \sum^{N}\mathbf{d_{L2}}(M,\Tilde{M})
\end{equation}
where $\mathbf{d_{L2}}$ means the $\mathbb{L}_2$ distance.
This procedure is pivotal in guiding the perception of instance embeddings. In the inference stage, the Instance Perception Decoder can be discarded.
The extracted image features $f^{Img}$ and instance embeddings $f^{Ins}$ are then sent to the Spatial-temporal Attention Module (SAM) to perform Joint Image-Instance Spatial-Temporal Attention.

\subsection{ Joint Image-Instance Spatial-Temporal Attention}
To generate video prototypes with improved discriminative capabilities, we propose a Spatial-temporal Attention Module (SAM) that effectively integrates image-level and instance-level information across spatial and temporal dimensions.
As depicted in Figure \ref{fig:framework}, the SAM consists of three primary components: spatial attention sub-module, temporal attention sub-module, and global fusion sub-module. 
We obtain spatial and temporal fusion features with the spatial and temporal attention sub-module, respectively.
After that, these features are summed and fed into the global fusion sub-module to obtain discriminative prototypes.

Specifically, in the spatial attention sub-module, we unfold the image features and instance embeddings along the spatial dimension ($f^{Img(S)},f^{Ins(S)} \in \mathbb{R}^{T\times HW \times C}$), apply 
 multi-head self-attention to each, and pass their sum through a feed-forward network to obtain the spatial fusion feature $f^{S}\in \mathbb{R}^{T\times HW \times C}$.
Similarly, within the temporal attention sub-module, the image features and instance embeddings are unfolded along the temporal dimension ($f^{Img(T)},f^{Ins(T)} \in \mathbb{R}^{HW\times T \times C}$), and produces the temporal fusion feature ${f^{T}}\in \mathbb{R}^{HW\times T \times C}$.
The process can be formulated as:
\begin{equation}
    {f^{i}} = \mathbf{FFN}(\mathbf{MHA}(f^{Img(i)})+\mathbf{MHA}(f^{Ins(i)}))
\end{equation}
where $i \in \{S,T\}$, $\mathbf{MHA}$ means multi-head self-attention and $\mathbf{FFN}$ refers to the feed-forward network with a residual connection.

Compared to previous work, this network design significantly enhances the network's capacity to integrate instance embeddings and image features across spatial and temporal dimensions, thereby enriching the feature representations with action-related instance information.

Subsequently, to merge the spatial and temporal features and construct robust video prototypes, we pool the spatial and temporal fusion features, sum them to obtain the spatial-temporal feature $f^{ST}$, and feed it into the global fusion sub-module.

In global fusion sub-module, we introduce a learnable token$f^{token} \in \mathbb{R}^C$ concatenated with the spatial-temporal feature $f^{ST}$ to adaptively aggregate global contextual information. The effectiveness of this approach has been demonstrated by multiple previous methods \citep{2023-molo,huang2022compound}. Additionally, positional embeddings are added to encode the relative positional relationships.
Finally, we employ multi-head attention to construct action prototypes.
The process can be formulated as:

\begin{equation}
\begin{aligned}
    &f^{ST} = \mathbf{AvgPool}(f^{S}) + \mathbf{AvgPool}(f^{T}) \\
    &\Tilde{f} = \mathbf{MHA}([f^{token}, f^{ST}] + f_{pos})
\end{aligned}
\end{equation}
where $\mathbf{AvgPool}$ means Average Pooling, $f^{token} \in \mathbb{R}^C$ is a learnable token, and $f_{pos} \in \mathbb{R}^{(T+1)\times C}$ represents the position embeddings used to encode the relative positional relationships. 
We take the output features $\Tilde{f} = \{ \Tilde{f}^{token},\Tilde{f}^1,\Tilde{f}^2,\ldots,\Tilde{f}^T\}$ as the action prototype. 
In this way, the output feature $\Tilde{f}^{token}$ corresponding to the learnable token has the global representation ability. And $\Tilde{f}^t$ denote the local features at each timestamp $t$, enriched with action-related instance information.  

We then show how we leverage the class property of the token feature for robust video prototype matching.


\begin{table*}[t]
\centering
\small
\caption{Comparison with recent state-of-the-art few-shot action recognition methods on the SSv2-Small and UCF101 datasets. We report the results on the 5-way 1shot, 3shot, and 5shot settings.
"-" indicates the result is not available in published works.
The best results are bolded and the underline means the second best performance.
}

\setlength{
    \tabcolsep}{
    2.0mm}{
\begin{tabular}
{l|c|c|ccc|ccc}
\hline
			
\hspace{-0.5mm}  \multirow{2}{*}{Method} \hspace{2mm} 
& \multicolumn{1}{c|}{\multirow{2}{*}{Reference}} 
& \multicolumn{1}{c|}{\multirow{2}{*}{Pre-training}} 
& \multicolumn{3}{c|}{{SSv2-Small}}  
& \multicolumn{3}{c}{{UCF101}}  \\
& \multicolumn{1}{c|}{} 
& \multicolumn{1}{c|}{} 
& \multicolumn{1}{l}{1-shot} & 3-shot  & 5-shot  
& \multicolumn{1}{l}{1-shot} & 3-shot  & 5-shot   \\ 
\shline

MatchingNet~\citep{vinyals2016matchingNet}    & NeurIPS'16 & ResNet-50
& 31.3  & 39.8  & 45.5   
& -   & -  & -  \\
Plain CMN~\citep{2018-CMN}    & ECCV'18 & ResNet-50
& 33.4  & 42.5  & 46.5     
& -   & -  & -   \\
CMN-J~\citep{2020-CMN}    & TPAMI'20 & ResNet-50
& 36.2  & 44.6  & 48.8   
& -   & -  & -    \\
ARN~\citep{ARN-ECCV}    & ECCV'20 & C3D
& -  & -  & -    
& 66.3   & -  & 83.1    \\
OTAM~\citep{2020-OTAM}    & CVPR'20 & ResNet-50
& 36.4  & 45.9  & 48.0     
& 79.9   & 87.0  & 88.9   \\
AMeFu-Net~\citep{AmeFu-Net} & ACM MM'20 & ResNet-50
& -  & -  & -     
& 85.1   & 93.1  & 95.5   \\
TRX ~\citep{perrett2021trx}  & CVPR'21    & ResNet-50      
&  36.0   &  51.9   &  {56.7}  
& 78.2  & {92.4}  & {96.1} \\  

TA$^{2}$N~\citep{TA2N}    & AAAI'22 & ResNet-50
& -  & -  & -  
& 81.9   & -  & 95.1   \\
STRM~\citep{STRM}    & CVPR'22  & ResNet-50
& 37.1 & 49.2 & 55.3 
& 80.5 & 92.7 & \underline{96.9}  \\
HyRSM~\citep{2022-HyRSM}    & CVPR'22  & ResNet-50
& 40.6  & 52.3  & 56.1   
& 83.9   & 93.0  & {94.7}  \\
CMPT~\citep{huang2022compound}    & ECCV'22   & ResNet-50
& 38.9  & -  & \underline{61.6}   
& 71.4   & -  & 91.0   \\

HCL~\citep{HCL}    & ECCV'22  & ResNet-50
& 38.7  & 49.1  & 55.4   
& 73.7   & 82.4  & 85.8  \\

MPRE~\citep{MPRE} & TCSVT'22  & ResNet-50
& -  & -  & -
& 82.0   & -  & {96.4}  \\
VIM~\citep{VIM} & arXiv'23  & ResNet-50
& 40.8  & - & 54.9 
& 86.2  & -  &  96.1 \\
SA-CT~\citep{SA-CT}    & ACM-MM'23  & ResNet-50
& -  & - & - 
& 85.4  & -  &  {96.4} \\

{GgHM} \citep{xing2023GgHM}  &{ICCV'23}  & {ResNet-50}
& {-}  & {-} & {-}
& {{85.2}}   & {-}  & {96.3} \\

MoLo \citep{2023-molo}  &CVPR'23  & ResNet-50
& {42.7}  & \underline{52.9}  & {56.4}
& {86.0}   & \underline{93.5}  & {95.5} \\

{MGTSN~\citep{mgtsn}} & {Nc'24}  & {ResNet-50}
& {-}  & {-} & {-} 
& {\underline{86.3}}  & {-}  &  {97.6} \\

HCR~\citep{hcr} & CVIU'24  & ResNet-50
& -  & - & - 
& 82.4  & -  &  93.2 \\

{HYRSM++~\citep{wang2024hyrsm++}} & {PR'24}  & {ResNet-50}
& {\underline{42.8}}  & {52.4} & {58.0}
& {85.8}  & {\underline{93.5}} &  {95.9} \\

{Huang \textit{et al.}~\citep{huang2024matching}} & {IJCV'24}  & {ResNet-50}
& {42.6}  & {-} & {\textbf{61.8}}
& {74.9}  & {-} &  {92.5} \\

\shline
Bi-MHM~\citep{2022-HyRSM}   & CVPR'22  & ResNet-50
& 38.0  & 47.6 & 48.9
& 81.7   & 88.2  & 89.3  \\
\rowcolor{Gray}
I$^2$ST  & -  & ResNet-50
& \textbf{43.6}     & \textbf{54.6}   & {57.6}  
& \textbf{87.7}   & \textbf{94.8}  & {95.7}   \\ 

\hline
\end{tabular}
}

\label{tab:compare_SOTA_1}
\end{table*}

\subsection{Global-Local Prototype Matching}
To address the challenge of robust video matching, existing methods \citep{2023-molo, wang2023clip} typically classify query video sample $\Tilde{f}_q$ with the frame-level matching metrics based on features from randomly sparse sampled frames.
However, such random sparse sampling may introduce excessive background information (e.g., consecutive background frames), leading to errors during frame-level metric matching.
To overcome this limitation, we propose Global-Local Prototype Matching, which incorporates global features to capture the action characteristics of the entire temporal sequence. By emphasizing the complete action sequence, this approach mitigates the issues caused by sparse frame sampling in frame-level matching metrics.
Therefore, in our framework, the similarity score $D_{i,q}$ is computed by the weighted average of global feature $\Tilde{f}^{token}$ and local feature$\{\Tilde{f}^1,\Tilde{f}^2,...,\Tilde{f}^T\}$. 
which can be expressed as:
\begin{equation}
\begin{aligned}
D_{i,q}& = \mathbf{M}([\Tilde{f}_i^1,...,\Tilde{f}_i^T],[\Tilde{f}_q^1,...,\Tilde{f}_q^T])  \\
&+\mathbf{Sim}(\Tilde{f}_i^{token},\Tilde{f}_q^{token})
\end{aligned}
\end{equation}
where $\mathbf{M}$ is a frame-level metric (Bi-MHM \citep{2022-HyRSM} in our framework), and $\mathbf{Sim}$ denotes the cosine similarity function. 
Then we can use the output support-query similarity scores $D_{i,q}$ as logits for classification.
Experiments demonstrate that the incorporation of global features contributes to improving the performance of robust video matching, as detailed in Section 4.

During the training phase, the entire framework is trained end-to-end, and the final loss can be denoted as:
\begin{equation}
    \mathcal{L} = \mathcal{L}_{CE} 
    + \lambda\mathcal{L}_{Mask} 
\end{equation}
where $\lambda$ is a balanced factor. $\mathcal{L}_{CE}$ is the cross-entropy loss over the support-query distances based on the ground-truth label.
$\mathcal{L}_{Mask}$ represents the $\mathbb{L}2$ loss of action-related instance masks for the Instance Perception module.

\section{Experiments}
In this section, we first introduce the experimental settings of our I$^2$ST in Section 4.1 and then compare our approach with previous state-of-the-art methods on five commonly used benchmarks in Section 4.2. 
Finally, comprehensive ablation studies of I$^2$ST are provided in Section 4.3, with the visualizations of the experiments presented in Section 4.4 to demonstrate the effectiveness of each module.

\subsection{Datasets and experimental setup}

\subsubsection{\textbf{Datasets.}}
We perform our experiments on five widely used Few-shot Action Recognition datasets, including SSv2-Full \citep{goyal2017ssv2}, SSv2-Small \citep{goyal2017ssv2}, Kinetics \citep{carreira2017kinetics}, UCF101 \citep{soomro2012ucf101}, and HMDB51 \citep{kuehne2011hmdb51}. 
For SSv2-Full and SSv2-Small, we follow the dataset split as described in \citep{2020-OTAM} and \citep{2018-CMN}. Specifically, 64 classes from the original dataset are allocated for training, while 24 classes are reserved for testing.
The key difference between SSv2-Full and SSv2-Small lies in the dataset size: SSv2-Full includes all available samples for each category, whereas SSv2-Small is limited to 100 samples per category.
The Kinetics dataset, employed in the few-shot setup \citep{2020-OTAM, 2022-HyRSM, 2023-molo}, is a subset of the original dataset \citep{carreira2017kinetics}. 
The UCF101 \citep{soomro2012ucf101} comprises 101 action classes, and we adhere to the few-shot split described in \citep{2022-HyRSM,MTFAN,ARN-ECCV}. 
For HMDB51 \citep{kuehne2011hmdb51}, we follow the standard practice of using 31 classes for training and 10 classes for testing.


%
\begin{table*}[t]
\centering
\small
\caption{Comparison with state-of-the-art few-shot action recognition methods on Kinetics, SSv2-Full, and HMDB51 in terms of 1-shot and 3-shot classification accuracy.
"-" stands for the result is not available in published works.
The best results are bolded in black, and the underline represents the second best result.
}

\setlength{
    \tabcolsep}{
    2.0mm}{
\begin{tabular}
{l|c|c|cc|cc|cc}
\hline
			
\hspace{-0.5mm}  \multirow{2}{*}{Method} \hspace{2mm} 
& \multicolumn{1}{c|}{\multirow{2}{*}{Reference}} 
& \multicolumn{1}{c|}{\multirow{2}{*}{Pre-training}} 
& \multicolumn{2}{c|}{{Kinetics}}  &\multicolumn{2}{c|}{{SSv2-Full}}  
& \multicolumn{2}{c}{{HMDB51}}  \\
& \multicolumn{1}{c|}{} 
& \multicolumn{1}{c|}{} 
& \multicolumn{1}{l}{1-shot} & 3-shot   
& \multicolumn{1}{l}{1-shot} & 3-shot   
& \multicolumn{1}{l}{1-shot}& 3-shot      \\ \shline
MatchingNet~\citep{vinyals2016matchingNet}    & NeurIPS'16  & ResNet-50
& 53.3  & 69.2  
& -   & -   
& -  & -     \\
Plain CMN~\citep{2018-CMN}    & ECCV'18  & ResNet-50
& 57.3  & 72.5  
& -   & -  
& -  & -     \\
CMN-J~\citep{2020-CMN}    & TPAMI'20  & ResNet-50
& 60.5  & 75.6   
& -   & -  
& -  & -    \\
ARN~\citep{ARN-ECCV}    & ECCV'20  & C3D
& 63.7  & -  
& -   & -    
& 45.5  & -      \\
OTAM~\citep{2020-OTAM}    & CVPR'20  & ResNet-50
& 73.0  & 78.7    
& 42.8  & 51.5   
& 54.5  & 65.7      \\
AMeFu-Net~\citep{AmeFu-Net} & ACM MM'20 & ResNet-50
& 74.1  & 84.3  
& -     & -   
& 60.2  & 71.5   \\
TRX~\citep{perrett2021trx}    & CVPR'21  & ResNet-50
& 63.6  & 81.1  
& 42.0  & 57.6 
& 53.1  & 66.8    \\
TA$^{2}$N~\citep{TA2N}    & AAAI'22  & ResNet-50
& 72.8  & - 
& 47.6  & -  
& 59.7  & -     \\
STRM~\citep{STRM}    & CVPR'22  & ResNet-50
& 62.9 & 81.1
& 43.1 & 59.1
& 52.3 & 67.4    \\
HyRSM~\citep{2022-HyRSM}    &  CVPR'22  & ResNet-50
& 73.7  & 83.5  
& 54.3  & {65.1} 
& 60.3  & 71.7    \\
CMPT~\citep{huang2022compound}    & ECCV'22  & ResNet-50
& 73.3  & -   
& 49.3  & -  
& 60.1  & -      \\ 
HCL~\citep{HCL}    & ECCV'22   & ResNet-50
& 73.7  & 82.4   
& 47.3  & 59.0  
& 59.1  & 71.2    \\ 
MPRE~\citep{MPRE} & TCSVT'22  & ResNet-50
& 70.2  & - 
& 42.1  & -
& 57.3  & -  \\
SloshNet~\citep{sloshnet} & arXiv'23   & ResNet-50
& {-}  & - 
& {46.5} & - 
& {-}  &  - \\

VIM~\citep{VIM} & arXiv'23   & ResNet-50
& {73.8}  & - 
& - & - 
& {61.1}  &  - \\
SA-CT~\citep{SA-CT}    & ACM-MM'23   & ResNet-50
& 71.9   & -
& 48.9  &  -
& 60.4  & -  \\
{GgHM~\citep{xing2023GgHM}}  &{ICCV'23}  & {ResNet-50}
& {\underline{74.9}}    & {-} 
& {54.5}  & {-}
& {61.2}  & {-} \\

MoLo~\citep{2023-molo}    & CVPR'23   & ResNet-50
& {74.0}   & {83.7} 
& \underline{56.6}  &  \underline{67.0}
& {60.8}  & \underline{72.0}   \\

TSA-MLT~\citep{tsa-mlt} & Nc'24   & ResNet-50
& {66.8}  & - 
& {43.8} & - 
& {57.9}  &  - \\

{MGTSN~\citep{mgtsn}} & {Nc'24}  & {ResNet-50}
& {73.9}  &  {-} 
& {51.6}  &  {-}
& {60.2}  &  {-} \\

HCR~\citep{hcr} & CVIU'24  & ResNet-50
& {70.2}  & - 
& -   &  -
& \underline{62.5} &  - \\

{HYRSM++~\citep{wang2024hyrsm++}} & {PR'24}  & {ResNet-50}
& {74.0}  & {\underline{83.9}}
& {55.0}   &  {66.0}
& {61.5} &  {72.7} \\

{Huang~\textit{et al.}~\citep{huang2024matching}} & {IJCV'24}  & {ResNet-50}
& {74.0}  & {-}
& {52.3}   &  {-}
& {61.6} &  {-} \\

\shline
Bi-MHM~\citep{2022-HyRSM}    & CVPR'22  & ResNet-50
& 72.3  & 81.1
& 44.6   & 53.1   
& 58.3  & 67.1  \\
\rowcolor{Gray}

I$^2$ST     & -   & ResNet-50
& \textbf{75.0}  & \textbf{84.0} 
& \textbf{57.6}  & \textbf{67.4}
& \textbf{63.1}  & \textbf{74.0}   \\

\hline
\end{tabular}
}
\label{tab:compare_SOTA_2}
\end{table*}

\subsubsection{\textbf{Implementation details.} }
For a fair comparison with existing methods \citep{2020-OTAM,perrett2021temporal,2018-CMN,2023-molo,2022-HyRSM}, our approach also employs a ResNet50 \citep{2016-resnet} pre-trained on ImageNet \citep{russakovsky2015imagenet} as the basic feature extractor and removes the last global-average pooling layer to retain spatial information. 
We optimize our I$^2$ST end-to-end with Adam optimizer \citep{kingma2014adam} and apply an auxiliary contrastive loss to stabilize the training process as in\citep{BAM_CML, HCL, wang2023clip, 2023-molo}.
Following previous methods \citep{2020-OTAM,perrett2021temporal,2022-HyRSM,2023-molo}, we uniformly and sparsely sample 8 frames (\textit{i.e.}, $T = 8$) to represent the entire video, and employ the Bi-MHM \citep{2022-HyRSM} as the frame-level metric.  
During the training phase, we employ standard data augmentation techniques, including random cropping and color jittering. For few-shot testing, we centrally crop a $224 \times 224$ region from each frame.
For many-shot inference, such as 3-shot and 5-shot, we adopt the averaged prototype paradigm \citep{snell2017prototypical} to classify query video samples. 
Following previous methods \citep{STRM,2022-HyRSM, 2023-molo}, we calculate the average classification accuracy of the 10,000 episode task from the test set to evaluate the few-shot performance on each benchmark.

\subsection{Comparison with state-of-the-art methods}
We compare the performance of I$^2$ST with state-of-the-art methods in this section. As shown in Table \ref{tab:compare_SOTA_1} and Table \ref{tab:compare_SOTA_2}, we can make the following observations:
(a) Compared with the Bi-MHM baseline, our method can significantly improve the performance by integrating the image-level information and instance-level information. For example, under the 5-way 1-shot setting, our method achieves 5.6\% and 6.0\% on the SSv2-Small and UCF101 datasets, respectively, which validates the effectiveness of our motivation to perceive action-related instances and integrate the instance-level and image-level features across spatial and temporal dimension.
(b) Our proposed I$^2$ST consistently outperforms other methods significantly across different datasets with 1-shot and 3-shot settings and achieves competitive results in the 5-shot setting.
Notably, compared to other foreground perception-based FASR methods(\textit{e.g.} HCL \citep{HCL}, SA-CT \citep{SA-CT}, SloshNet \citep{sloshnet},CMPT \citep{huang2022compound}, Huang \textit{et al.} \citep{huang2024matching}), our method achieves better performance in most settings, demonstrating that our approach can robustly extract action features and reduce the impact of background noise in few-shot scenarios.
(c) Experiments show that our method significantly improves performance in few-shot scenarios, with particularly notable improvements in the 1-shot setup compared to the 3-shot and 5-shot settings.
We believe this is because, in the 1-shot scenario, the model is more easily influenced by background features, making it difficult to learn the intrinsic features of the action, leading to misclassifications. As the number of samples in the support set increases, the model can learn robust action features across different backgrounds, thereby limiting the performance improvement provided by the additional instance information. It suggests that our method can learn rich and effective representations with extremely limited samples.

It is noteworthy that the SSv2-Small and SSv2-Full datasets contain a large number of actions that require temporal understanding, such as "Picking [something] Up". While the Kinetics and UCF101 datasets are appearance-related datasets, which contain a wider variety of appearance-based action scenes, such as "Cutting Watermelon", and the HMDB51 dataset is relatively complicated and might involve diverse object interactions.
The excellent performance on these datasets shows that our proposed I$^2$ST has a strong generalization ability across different dataset scenarios.

\subsection{Ablation study}
For ease of comparison, we use a baseline method Bi-MHM \citep{2022-HyRSM} that applies global-average pooling to backbone representations to obtain a prototype for each class and perform Bi-MHM for video matching. We will explore the role and validity of our proposed modules in detail below.

\begin{table}[t]
\centering
\small
\caption{Ablation study on HMDB51 datasets under 5-way 1-shot
and 5-way 3-shot settings. The top line represents the baseline Bi-MHM. "T-Branch" and "S-Branch" denotes the temporal fusion branch and  spatial fusion branch, respectively.
}
\setlength{
    \tabcolsep}{
    0.6mm}{
\begin{tabular}
{ccc|cc}

\hline
			
\hspace{-0.8mm}  \multirow{3}{*}{Instance Perception} \hspace{0.5mm} & 
\hspace{-0.8mm}  \multirow{3}{*}{T-Branch} \hspace{0.5mm} &
\hspace{-0.8mm}  \multirow{3}{*}{S-Branch} \hspace{0.5mm} &

\multicolumn{2}{c}{{HMDB51}}    \\
 & & & \multicolumn{1}{l}{1-shot} & \multicolumn{1}{l}{3-shot}   \\ 
\shline

& & &                                  {58.3} & {67.1}     \\ 
\hline
&\checkmark & &                        60.8 & 72.5     \\ 
& & \checkmark &                       60.4 & 72.6     \\ 
& \checkmark & \checkmark &            61.1 & 72.5     \\ 
\hline
%
%
\checkmark &  &   &                     57.5 & 70.9   \\ 
\checkmark & \checkmark &   &           61.8 & 72.4   \\ 
\checkmark &  & \checkmark &            61.1 & 72.7   \\ 
\checkmark & \checkmark &  \checkmark & 63.1 & 74.0   \\ 

\hline

\end{tabular}
}
\label{tab:network_component_ablation}
\end{table}

\subsubsection{\textbf{Analysis of network components.}}
To investigate the effectiveness of different components in I$^2$ST, we selected Bi-MHM \citep{2022-HyRSM} as the baseline and performed ablation studies in a 5-way setting to thoroughly investigate the roles and effectiveness of our proposed modules, as detailed below.
We provide a detailed ablation analysis of the model components, with the experimental results summarized in Table \ref{tab:network_component_ablation}. 
Based on the results, we can observe that the Instance Perception Module and Spatial-temporal Attention Module play a key role in improving performance. 
In particular, 1-shot performance can be improved from 61.1\% to 63.1\% by adding the Instance Perception Module to the baseline, reflecting the importance of action-related instances in the process of video prototype learning.
The performance gain is also significant by introducing the Spatial-temporal Attention Module. 
In particular, the combination of the temporal and spatial branches yields better results than using a single branch alone. Directly incorporating Instance Perception into the BI-MHM method leads to a noticeable drop in performance in the 1-shot setting, highlighting the importance of our Spatial-temporal Attention Module.
Moreover, compared to the 3-shot setting, I$^2$ST's components show more significant improvement in the 1-shot scenario, consistent with our observations on other datasets.
Additionally, the experimental results demonstrate the complementarity between different modules, highlighting the rationality of our model's structure. 
By integrating all modules, our method improves performance to 63.1\% in 1-shot and 74.0\% in 3-shot settings, up from 58.3\% and 67.1\% compared to Bi-MHM, respectively.

Furthermore, in our approach, we utilize SEEM \citep{SEEM} to explicitly guide the learning of the Instance Perception Module. The scale and diversity of SEEM's training dataset, along with its open-vocabulary capabilities, help reduce the risk of bias towards specific text inputs, ensuring robustness across diverse scenarios.
Although the SEEM model may exhibit slight biases in certain categories within the Action Recognition context due to dataset differences, our approach mitigates this by adaptively integrating image-level and instance-level features across both spatial and temporal dimensions.
This fusion strategy reduces reliance on instance embeddings and allowing the model to better adapt to potential biases or variability, ensuring more robust performance.

%
\begin{table}[t]
\centering
\small
\caption{Comparison experiments on the effect of instance-level information, and different fusion architecture on the SSv2-Small and UCF101 datasets. 'I$^2$ST(cross)' means the cross-attention based framework, and 'I$^2$ST(self)' means the self-attention based framework.
}

\setlength{
    \tabcolsep}{
    0.8mm}{
\begin{tabular}
{c|cc|cc}
\hline
			
\hspace{-0.8mm}  \multirow{2}{*}{Setting} \hspace{0.5mm} & 
 \multicolumn{2}{c|}{{SSv2-small}} & \multicolumn{2}{c}{{UCF101}}    \\

& \multicolumn{1}{c}{1-shot} & 3-shot  
& \multicolumn{1}{c}{1-shot} & 3-shot \\ \shline

\hspace{-1mm} OTAM 
& 36.4 & 45.9
& 79.9 & 87.0  \\ 

\hspace{-1mm} OTAM+IPM
& 39.0\color{blue}{(+2.6)} & 47.5\color{blue}{(+1.6)}
& 82.5\color{blue}{(+2.6)}  &  87.9\color{blue}{(+0.9)}  \\ 

\hline

\rowcolor{Gray}
\hspace{-1mm} I$^2$ST(cross)
& 44.1  & 53.4    
& 87.6  & 94.5  \\ 

\rowcolor{Gray}
\hspace{-1mm} I$^2$ST(self)
& 43.6 & 54.6    
& 87.7  & 94.8  \\

\hline
\end{tabular}
}

\label{tab:ablation_archi}
\end{table}


%
\begin{table}[t]
\centering
\small
\caption{Comparison experiments on the effect of video match mechanism on the SSv2-Small, HMDB51 and UCF101 datasets. "L" means local prototype matching strategy and "G+L" means our Global-Local Prototype Matching strategy.
}

\setlength{
    \tabcolsep}{
    1.0mm}{
\begin{tabular}
{c|cc|cc|cc}
\hline
			
\hspace{-0.8mm}  \multirow{2}{*}{Setting} \hspace{0.5mm} & 
\multicolumn{2}{c|}{{SSv2-Small}}  & \multicolumn{2}{c|}{{HMDB51}} &\multicolumn{2}{c}{{UCF101}}  \\

& \multicolumn{1}{l}{1-shot} & 3-shot  
& \multicolumn{1}{l}{1-shot} & 3-shot
& \multicolumn{1}{l}{1-shot} & 3-shot \\ \shline

OTAM(L)
& 41.7 & 53.7
& 61.9  & 72.5
& 86.3  &  94.4  \\ 
Bi-MHM(L) 
& 41.8   & 54.0
& 62.2   & 72.5    
& 86.5   & 94.2  \\ 

\hline
\rowcolor{Gray}
I$^2$ST(G+L)
& \textbf{42.5}  & \textbf{54.6}
& \textbf{63.1}  & \textbf{74.0}    
& \textbf{87.1}  & \textbf{94.8}  \\ 

\hline
\end{tabular}
}

\label{tab:ablation_matching}
\end{table}


\subsubsection{\textbf{Effectiveness of the Instance-level information}}
To better demonstrate the improvements brought by instance-level information to Few-shot Action Recognition tasks, we have integrated the Instance Perception Module into the existing Few-shot Action Recognition method to incorporate instance-level information.
As shown in Table \ref{tab:ablation_archi}, 'OTAM+IPM' represents the introduction of instance-level information into the original 'OTAM' methods.
It can be observed that after incorporating instance-level information, OTAM methods achieve significant performance improvements on the SSv2-Small and UCF101 datasets under the 1-shot and 3-shot settings. 
This suggests that introducing instance-level information is effective in Few-shot Action Recognition, especially with extremely limited support set samples.

Moreover, to explore better ways of integrating image-level information with instance-level information, we have designed different fusion architectures of the Saptial/Temporal Attention Sub-module: Self-attention based and Cross-attention based. 
We present the self-attention based framework, where the Spatial/Temporal Attention Sub-module merges the image-level information and instance-level information using a self-attention mechanism.
In the Cross-attention based framework, we replace the two multi-head Attention modules with a cross-attention module, mapping instance embeddings as the query vector, while using image features as the key and value vectors.
As shown in Table \ref{tab:ablation_archi}, both of them achieved state-of-the-art performance. 
Compared to the Cross-attention based framework, the Self-attention based framework showed better performance. We attribute this to the differences between the instance embeddings obtained from IPM and the image features in encoding foreground instances, resulting in limited performance improvements with the cross-attention framework.
Therefore, we ultimately adopted the Self-attention based framework.

\subsubsection{\textbf{Analysis of the video match Mechanism.}}
In Table \ref{tab:ablation_matching}, we explore the role of Global-Local Prototype Matching in the proposed I$^2$ST framework. It shows that in our framework, the Bi-MHM matching method offers a slight improvement over OTAM, and both of them belong to the local prototype matching strategy.
However, with the Global-Local Prototype Matching, I$^2$ST achieved 1.2\% and 1.5\% improvements in 1-shot and 3-shot scenarios, respectively, on HMDB51. Additionally, improvements of varying degrees have been observed across different settings in other datasets.
This consistent improvement underscores the importance of Global-Local Prototype Matching for FSAR. 
We observe that the performance gain is particularly significant in the 1-shot scenario, suggesting that it would be more effective to leverage global information when reference sample videos are scarce.

\begin{table*}[ht]
\centering
\small
\caption{$N$-way 1-shot classification accuracy comparison with recent few-shot action recognition methods on the test sets of SSv2-Full and Kinetics datasets. 
The experimental results are reported as the way increases from 5 to 10.
}
\setlength{
    \tabcolsep}{
    1.4mm}{
\begin{tabular}
{c|cccccc|cccccc}
\hline
			
\hspace{-0.5mm}  \multirow{2}{*}{Method} \hspace{2mm} &
\multicolumn{6}{c|}{{SSv2-Full}}  & \multicolumn{6}{c}{{Kinetics}}  \\

& \multicolumn{1}{l}{5-way} & 6-way  & 7-way  & 8-way   & 9-way & 10-way 
& \multicolumn{1}{l}{5-way} & 6-way  & 7-way  & 8-way   & 9-way & 10-way   \\ \shline

OTAM\citep{2020-OTAM}    & 42.8 
& 38.6  & 35.1  & 32.3   & 30.0   & 28.2   
& 72.2   & 68.7  & 66.0  & 63.0  &  61.9 & 59.0 \\
TRX\citep{perrett2021trx}  & 42.0&      41.5    & 36.1  &  33.6   &  32.0   &  30.3   &  63.6   & 59.4  & 56.7  & 54.6  & 53.2  & 51.1  \\  
HyRSM\citep{2022-HyRSM}   
& 54.3     &  50.1    & {45.8}  & {44.3}  & {42.1}  & {40.0}   
& {73.7}   & {69.5}   & {66.6}  & {65.5}  & {63.4}  & {61.0}  \\
MoLo\citep{2023-molo}  
& \underline{56.6} & \underline{51.6} & \underline{48.1}  & \underline{44.8}  & \underline{42.5}   & \underline{40.3}   
& \underline{74.0}  &  \underline{69.7}   & \underline{67.4}  & \underline{65.8}  &  \underline{63.5}  &  \underline{61.3}  \\ 
\hline

\rowcolor{Gray}
I$^2$ST 
& \textbf{57.6} & \textbf{52.7} & \textbf{49.0}  & \textbf{46.0}  & \textbf{42.7}   & \textbf{41.2}   
& \textbf{75.0}  &  \textbf{71.4}   & \textbf{69.6}  & \textbf{66.5}  &  \textbf{64.4}  &  \textbf{62.3}  \\ 
\hline
\end{tabular}
}
\label{tab:compare_Nway}
\end{table*}

\subsubsection{\textbf{Analysis of $N$-way classification}}
We also analyze the effect of varying $N$ on the few-shot performance. 
We further conduct ablation experiments to show the $N$-way $1$-shot accuracy, where $N$ ranges from $5$ to $10$.
In Table \ref{tab:compare_Nway}, we compare I$^2$ST with existing methods, including OTAM, TRX, HyRSM, and MoLo.
It can be observed that as $N$ increases, the classification becomes more challenging, leading to a noticeable decline in performance. 
%
Specifically, the 10-way 1-shot accuracy of I$^2$ST drops by 12.7\% compared to the 5-way 1-shot result on Kinetics (75.0\% vs. 62.3\%). 
Notably, our method consistently outperforms others in various settings on both SSv2-Full and Kinetics datasets, which demonstrates the effectiveness and generalizability of I$^2$ST across different datasets and settings.

\begin{figure}[t]
  \centering
   \includegraphics[width=0.95\linewidth]{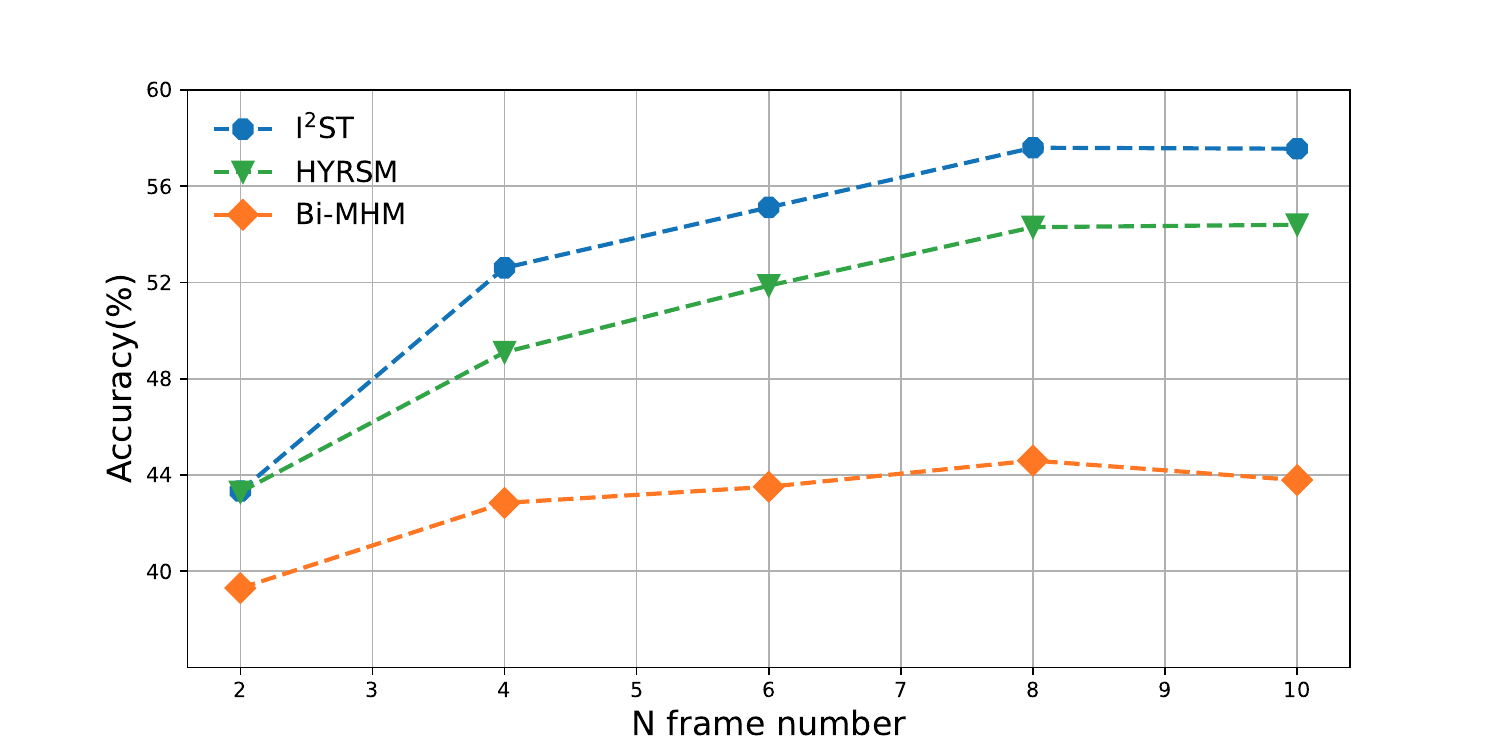}
   \caption{Ablation study on the effect of changing the number of input video frames under the 5-way 1-shot SSv2-Full setting.}
   \label{fig:n_frames}
\end{figure}

\subsubsection{\textbf{Influence of frame numbers}}
For a fair comparison, our I$^2$ST is compared with the current methods under the condition of 8 frames of input. To analyze the impact of the number of input frames on the few-shot performance, we conduct experiments with input video sampling from 2 to 10 frames.
As shown in Figure \ref{fig:n_frames}, the performance starts to rise and gradually saturates as the number of input frames increases. Notably, our I$^2$ST consistently maintains leading performance, even being competitive with the results of Bi-MHM \citep{2022-HyRSM} using 8 frames of input when only 2 frames are used, indicating the effectiveness and robustness of our method.

%
\begin{table}[t]
\centering
\small
\caption{Performance comparison under crossdataset setting. We train the following models on the Kinetics dataset and evaluate them on HMDB51 and SSv2-small datasets. The bolded means the  best performance. 
}

\setlength{
    \tabcolsep}{
    1.0mm}{
\begin{tabular}
{c|cc|cc}
\hline
			
\hspace{-0.8mm}  \multirow{2}{*}{Setting} \hspace{0.5mm} & 
 \multicolumn{2}{c|}{{HMDB51}} & \multicolumn{2}{c}{{SSv2-Small}}    \\

& \multicolumn{1}{c}{1-shot} & 3-shot  
& \multicolumn{1}{c}{1-shot} & 3-shot \\ \shline

OTAM \citep{2020-OTAM}
& 55.5& 62.8
&30.1 &34.7  \\ 

Bi-MHM \citep{2022-HyRSM}
&55.5 & 62.6
&30.4 & 35.1  \\ 

MoLo \citep{2023-molo}
& 54.4   & 67.7
& 28.6   &  36.4  \\ 

\hline
\rowcolor{Gray}
 I$^2$ST
& \textbf{56.2} & \textbf{68.8}   
& \textbf{30.9}  & \textbf{38.0}  \\ 

\hline
\end{tabular}
}

\label{tab:ablation_crossdataset}
\end{table}


\subsubsection{\textbf{Analysis of Cross-dataset setting}}
To further evaluate the generalization capability of our methods to unseen domains, in this section, we extend our evaluation under a cross-dataset setting. Specifically, we trained the models on the source dataset (Kinetics) and evaluated them on different target datasets (HMDB51 and SSv2-Small).
Compared with appearance-related dataset Kinetics, the HMDB51 and SSv2-small datasets contain more temporal action and might involve various object interactions.
As shown in Table \ref{tab:ablation_crossdataset}, I$^2$ST achieves the best performance on these totally different datasets compared with previous methods, demonstrating the effectiveness and strong generalizability of our methods. 

\begin{table}[t]
\centering
\small
\caption{Complexity analysis for 5-way 1-shot HMDB51 evaluation. The experiments are carried out on one Nvidia RTX 3090 GPU. ”I$^2$ST-S” and ”I$^2$ST-T” denotes the I$^2$ST only with the spatial fusion branch and temporal fusion branch, respectively.
}
\renewcommand{\arraystretch}{1.1} 
\setlength{
    \tabcolsep}{
    0.6mm}{ 
\begin{tabular}
{c|ccc|c}
\hline

Method & Param  & FLOPs & {Latency} & Acc\\

\hline

OTAM \citep{2020-OTAM} 
&23.5M & 8.2G &  {73.6ms} & 54.5 \\ 

OTAM(ResNet-101) 
&42.5M & 15.0G & {114.8ms}  & 59.9 \\ 

OTAM(ResNet-152)
&58.1M & 22.1G & {165.4ms} & 56.8 \\ 

TRX \citep{perrett2021trx}
&47.1M & 8.2G & {65.9ms} &53.1  \\ 
 
HYRSM \citep{2022-HyRSM} 
&65.6M & 8.4G & {64.1ms} &60.3  \\ 

MoLo \citep{2023-molo}
&89.6M & 10.1G & {108.2ms} &60.8  \\ %

\hline
\rowcolor{Gray}
I$^2$ST-S 
&88.5M & 13.4G  & {92.4ms} &61.1  \\
\rowcolor{Gray}
I$^2$ST-T 
&88.5M & 13.4G  & {92.5ms} &61.8  \\
\rowcolor{Gray}
I$^2$ST  
&130.2M & 17.1G & {114.3ms} &63.1  \\
\hline
\end{tabular}
}
\label{tab:ablation_complex}
\end{table}


\subsubsection{\textbf{Analysis of computational complexity}}

In order to further understand I$^2$ST, Table \ref{tab:ablation_complex} illustrates its differences with other methods in terms of parameters, computation and inference latency. As shown in Table \ref{tab:ablation_complex}, our method achieves better performance at the cost of a moderate increase in parameters and computational complexity. To explore the impact of the number of parameter, we also extended the OTAM method to the ResNet-101 and ResNet-152 backbones. Compared to the original OTAM \citep{2020-OTAM}, OTAM(ResNet-101) achieved a 5.1\% performance improvement, while OTAM(ResNet-152) only achieved a 2.3\% improvement, but introduced a significant increase in inference latency. This suggests that in few-shot scenarios, simply increasing network parameters does not necessarily result in effective performance gains and may instead lead to overfitting, which ultimately degrades performance. 
In contrast, our proposed I$^2$ST method demonstrates a more balanced trade-off. When using only the temporal or spatial branch, I$^2$ST achieves parameter efficiency comparable to MoLo \citep{2023-molo}, while providing better performance and faster inference speed. When the spatial and temporal branches are combined, I$^2$ST increases computational complexity but also provides further performance improvements. Thus, we believe that this level of complexity is justified by the significant performance gains it offers.

%
\begin{table}[t]
\centering
\small

\caption{{Performance comparison on video clips of varying lengths, conducted on the SSv2-Full dataset. "SSv2-3s" refers to video clips lasting 3 seconds or less, while "SSv2-5s" refers to those lasting 5 seconds or less. The best results are highlighted in bold.}
}

\renewcommand{\arraystretch}{1.1} 
\setlength{
    \tabcolsep}{
    1.4mm}{ 
\begin{tabular}
{c|ccc}
\hline

Method &  SSv2-3s  &  SSv2-5s & SSv2-Full \\
\hline

Bi-MHM \citep{2022-HyRSM} 
& 39.8 & 43.9 &  44.6 \\ 

MoLo \citep{2023-molo}
& 49.8 & 52.3 & 56.6   \\ %

\hline

I$^2$ST  
&\textbf{53.0} & \textbf{54.9} & \textbf{57.6}  \\
\hline

\end{tabular}
}

\label{tab:ablation_shortclip}
\end{table}


\subsubsection{\textbf{Analysis of short action clips}}

To further explore the generalizability  of our method in challenging scenarios, we compared its performance with other methods on videos of varying temporal lengths. From the SSv2-Full test dataset, we constructed two subsets: "SSv2-3s", containing video clips of 3 seconds or less, and "SSv2-5s", containing video clips of 5 seconds or less.
As shown in the table, our method outperforms both the baseline (Bi-MHM) and the SOTA method (MoLo) on the "SSv2-3s" and "SSv2-5s" subsets, demonstrating its effectiveness. Furthermore, compared to the SSv2-Full (Default) set, MoLo's performance drops by 6.8\% on the "SSv2-3s" subset, while our method shows a smaller drop of only 4.2\%. This highlights the superior generalization capability of our method on very short clips.
We attribute this superior performance to our method's ability to accurately capture action-related instances while minimizing the influence of action-irrelevant background information. Short clips often occur in relatively static environments, where sampled frames typically contain consistent background information. Excessive background redundancy can hinder the construction of accurate action prototype features. However, our method effectively captures action-related instance information, ensuring robustness and effectiveness in short clip scenarios.

\subsection{Visualization analysis}

\begin{figure}[t]
  \centering
   \includegraphics[width=0.95\linewidth]{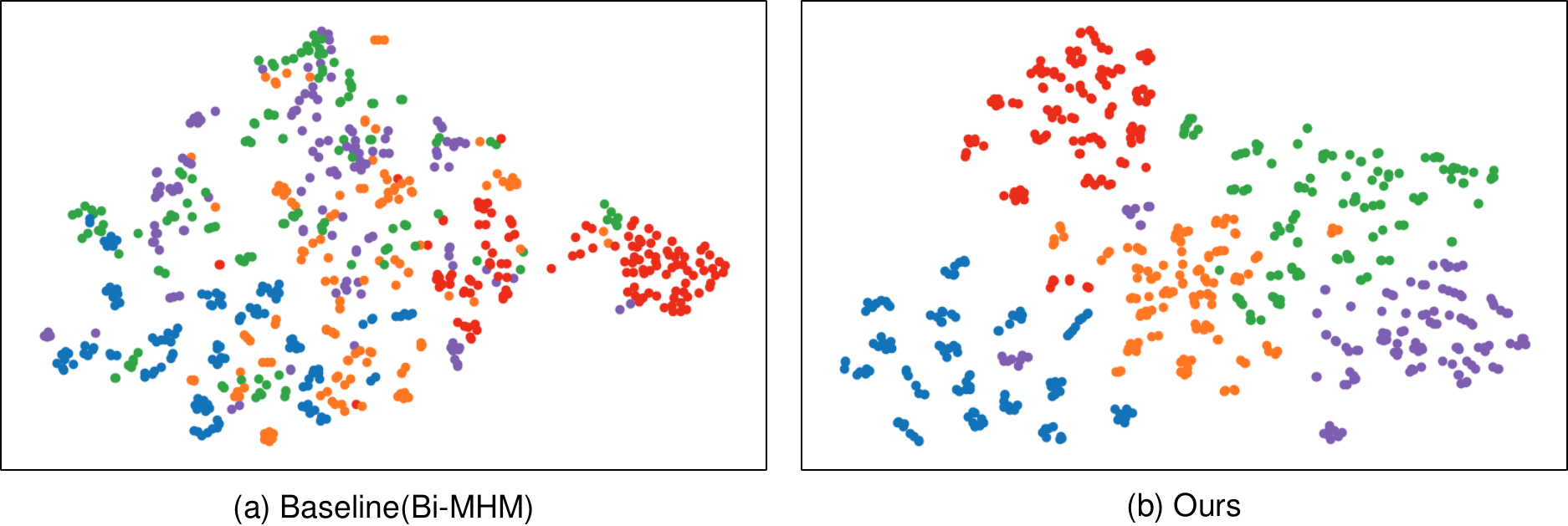}
   \caption{T-SNE distribution visualization of five action classes on test set of SSv2-Full. The different color represents video from different categories.}
   \label{fig:distribution_visualization}
\end{figure}

\begin{figure*}[t]
  \centering
   \includegraphics[width=0.95\linewidth]{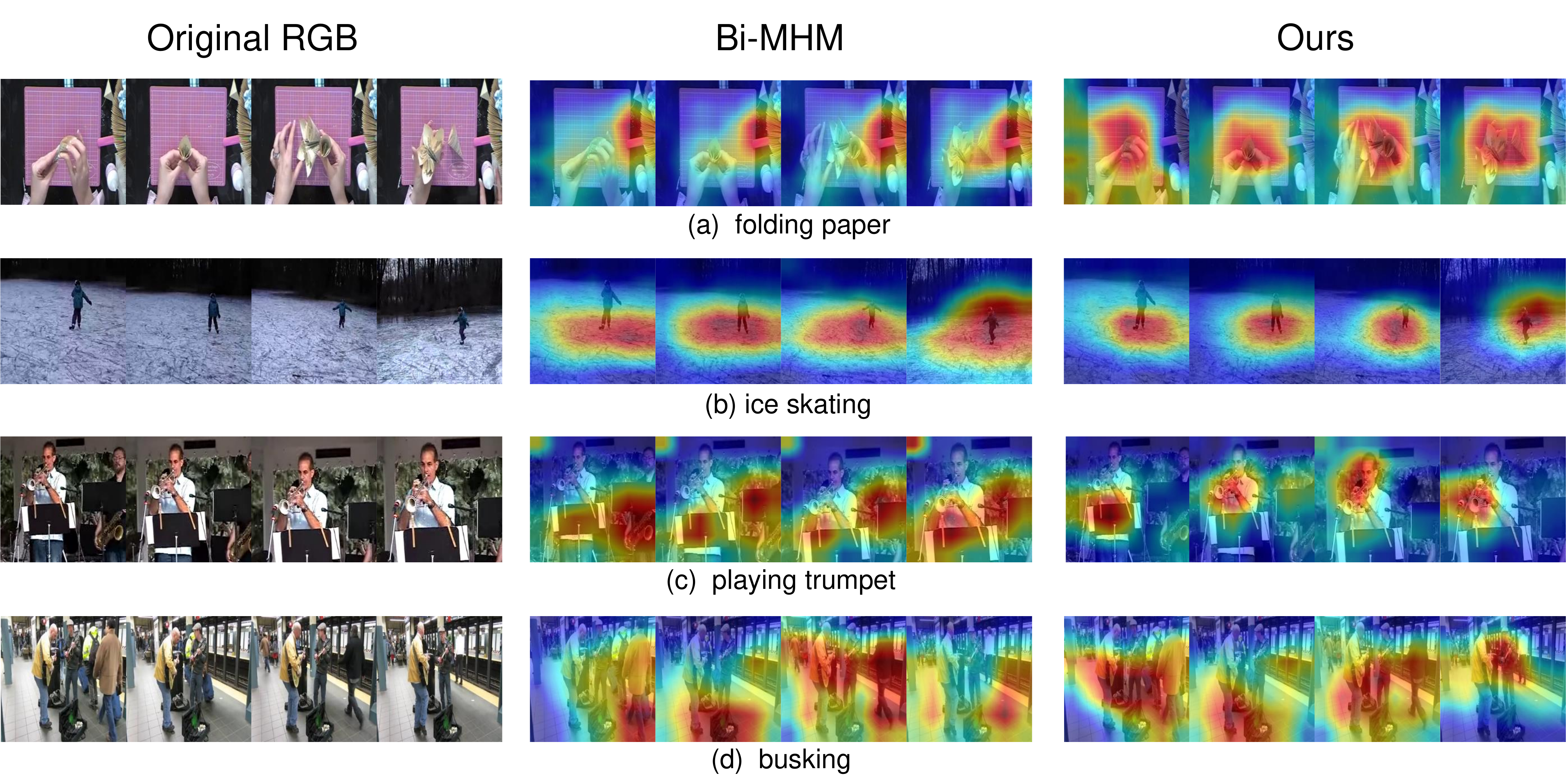}
   \caption{Visualization results of GradCAM \citep{selvaraju2017grad} heat maps on test sets of Kinetics dataset.}
   \label{fig:CAM_visualization}
\end{figure*}

\subsubsection{\textbf{Visualization of Prototype Similarity}}
To further qualitatively analyze our I$^2$ST, we visualize the feature distribution of the baseline methods and our method in the test phase, as shown in Figure \ref{fig:distribution_visualization}.
As shown in Figure \ref{fig:distribution_visualization}, with the Instance Perception Module and Spatial-temporal Attention Module, I$^2$ST significantly enhanced discriminative capabilities of the action video feature representations over the baseline Bi-MHM \citep{2022-HyRSM}, leading to improved prediction accuracy.

\subsubsection{\textbf{CAM Visualization}}

To further visually evaluate the proposed I$^2$ST, we compare its GradCAM\citep{selvaraju2017grad} visualization results with the baseline Bi-MHM \citep{2022-HyRSM} on Kinetics test set. 
As shown in Figure \ref{fig:CAM_visualization} our method focuses more precisely on action-related regions. In particular, as shown in Figures \ref{fig:CAM_visualization} (a) and \ref{fig:CAM_visualization} (b), our approach focuses more precisely on the regions where the action occurs in the video, while reducing the influence of background noise unrelated to the action. Furthermore, as shown in Figures \ref{fig:CAM_visualization} (c) and \ref{fig:CAM_visualization} (d), when the action takes place in a complex scene, our method can effectively focus on action-related instances, whereas Bi-MHM \citep{2022-HyRSM} is influenced by unrelated instances (e.g., irrelevant objects or people).

\begin{figure}[t]
  \centering
   \includegraphics[width=0.95\linewidth]{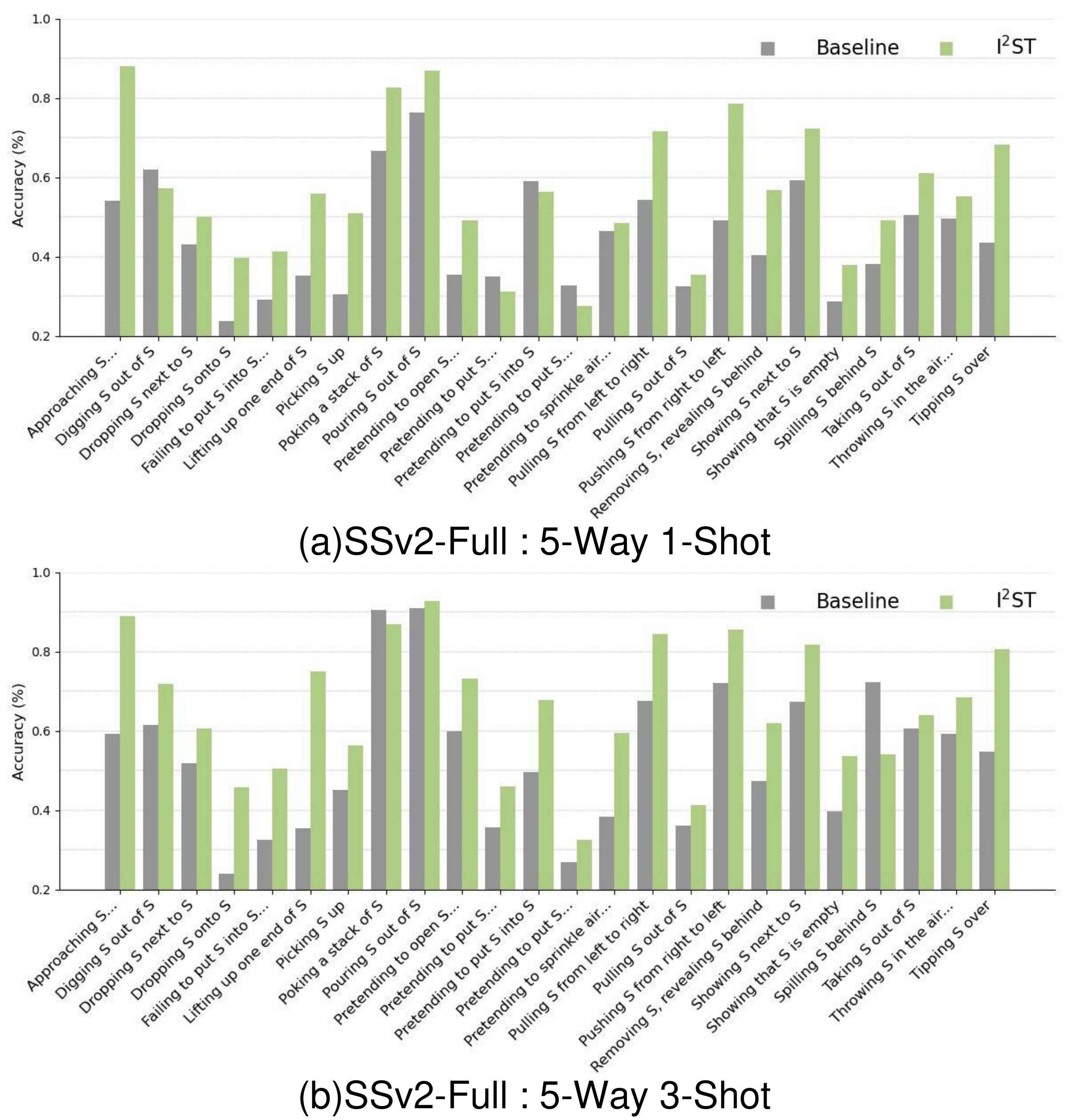}
   \caption{Quantitative analysis of 5-way 1-shot/3-shot class improvements on the SSv2-Full dataset. "S" is the abbreviation of "something.
   }
   \label{fig:ssv2visualization}
\end{figure}

\section{Limitations}

\textbf{(1) Failure Case:} To analyze the impact of the proposed method on the improvements of specific action classes, we perform statistics on the category performance gains compared to the baseline Bi-MHM \citep{2022-HyRSM} in Figure \ref{fig:ssv2visualization}.
We observe that most categories have a noticeable performance improvement over the baseline in SSv2-Full datasets with 5-way 1-shot/3-shot settings. However, there is a slight decrease in the performance of our method compared to the baseline on 1-shot setting in three categories: "Pretending to put [something] behind [something]", "Pretending to put [something] into [something]" and "Pretending to put [something] underneath [something]". These three action categories are related to "pretending", which leads to more complex temporal relationships, making them more prone to misclassification.

We attribute the performance drop to the insufficient modeling of long-term temporal sequences, especially in the 1-shot scenario. 
For example, in the action "pretending to put [something] into [something]", the model should focus on the entire action sequence, from putting the object in to taking it back. However, our method tends to focus more on individual action instance rather than modeling the relationships between actions across time. As a result, for actions like "pretend", which require long-term temporal modeling, the performance is weaker. 
In the feature, we will continue to explore more effective approaches for long-term sequence modeling to further improve our model's performance.

\textbf{(2) Computational Complexity:} 
Compared to CMPT \citep{huang2022compound}, our method dispenses with additional detector models in the testing phase, which significantly reduces model deployment complexity in practical applications. 
However, as shown in Table \ref{tab:ablation_complex}, while I$^2$ST achieves better performance across different datasets, it also introduces additional parameters (\textit{i.e.}, Spatial/Temporal Attention Module) into the model architecture, resulting in increased GPU memory and computation. 
We will continue to investigate how to reduce complexity without sacrificing performance in the future.

\section{Conclusion}
In this paper, we propose a novel I$^2$ST method for Few-shot Action Recognition, consisting of Action-related Instance Perception and Joint Image-Instance Spatial-Temporal Attention. 
The Action-related Instance Perception captures foreground action-related instance information to enhance the video representation, which is supervised by a text-based segmentation model. 
The Joint Image-Instance Spatial-Temporal Attention comprehensively integrates instance-level and image-level information across spatial and temporal dimensions, effectively leveraging both instance-level and image-level information to enhance the prototype representations.
Moreover, the proposed Global-Local Prototype Matching strategy achieves the robust and comprehensive few-shot action video matching with the global-local prototype.
In this way, our I$^2$ST can effectively exploit the foreground instance-level cues in complex few-shot video recognition scenarios, and improve the accuracy of few-shot action recognition.
Extensive experiments on five commonly used benchmarks verify the effectiveness of our method and show that I$^2$ST achieves state-of-the-art performance.

\section{Acknowledgments}
This work is supported by the STCSM under grant 22DZ2229005 and Shanghai Municipal Science and Technology Major Project under grant 2021SHZDZX0102.

\printcredits


\bibliographystyle{cas-model2-names}

\bibliography{cas-refs}



\end{document}